# Design of English-Hindi Translation Memory for Efficient Translation

Nisheeth Joshi
Department of Computer Science
Apaji Institute
Banasthali University
Rajasthan, India
nisheeth.joshi@rediffmail.com

Iti Mathur
Department of Computer Science
Apaji Institute
Banasthali University
Rajasthan, India
mathur_iti@rediffmail.com

**ABSTACT**

*Developing parallel corpora is an important and a difficult activity for Machine Translation. This requires manual annotation by Human Translators. Translating same text again is a useless activity. There are tools available to implement this for European Languages, but no such tool is available for Indian Languages. In this paper we present a tool for Indian Languages which not only provides automatic translations of the previously available translation but also provides multiple translations, in cases where a sentence has multiple translations, in ranked list of suggestive translations for a sentence. Moreover this tool also lets translators have global and local saving options of their work, so that they may share it with others, which further lightens the task.*

## 1. INTRODUCTION

Since the very genesis of the languages, translation has been the most important activity in day to day life. During ancient times traders and invaders, when traveled to foreign countries, had to learn new language or needed to hire a translator, who could translate text between their language and foreign language. In modern times, for the past sixty years, we have been trying to build systems which can automate the task of translation, These systems are popularly known as Machine Translators (MT) and their translations are termed as Machine Translation. The concept was first given by Bar Hillel [1] in 1951, where he mentioned about Fully Automatic, High Quality Translation (FAHQT), which more or less is still a distant dream.

In the last 15-20 years, with the incorporation of data driven approaches, there has been a paradigm shift. Two new approaches emerged in the arena of MT. These were Statistical Machine Translation (SMT) and Example Based Machine Translation (EBMT). Both relied heavily on the great amount of data for translation. But still, we are far from reaching our goal of FAHQT. One possible solution to this was proposed in the literature as Machine Aided Translation (MAT) or Computer Aided Translation (CAT) or as popularly known as Machine Aided Human Translation (MAHT). In this method, a translation is provided by the machine, since the output may not be of high quality, a human translator is required to edit translated text; this process is fairly common in today's MT scenario and is known as post editing.

Although machine translation with a reasonable accuracy is possible, it requires some existing data to be available for transition. This data is a comparative translation of two languages. One is called the Source Language (whose text is required to be translated) and the other is called Target Language (whose text is provided after translation). This text is termed as Bilingual Text or Parallel Text. In order to create such text we need human translators, who can provide correct translations for a given source text. One feature of a text, mostly found in technical documents, is the degree to which they resemble to each other. A user manual of a product is provided in different languages and has more or less same text as in other manuals of products in same category. One way to translate this repeated text is to copy the text from one location and paste it. But, then again this is a very tedious task, as looking back in a document for a similar looking phrase is very time consuming and taxing for the translator. What if he has to search other documents also, the problem become even more complex and time consuming. Using Computers we can solve this problem, we can have a translation aid which has a database of previously translated text fragments, which efficiently search for the correct translation. This translation is a draft or suggestive translation which can be modified by the translator. This transition aid is popularly known as Translation Memory (TM).

Through figure 1, it is clearly visible that, if we have greater no. of reparations or a large document then the performance of TM will increase. Thus giving us consistent and accelerated translations. This is fairly useful for large documentation projects where we constantly have to translate text in other languages. In the long run this reduces the cost of per sentence translation, as most of the fragments can be captured by the translation memory itself.

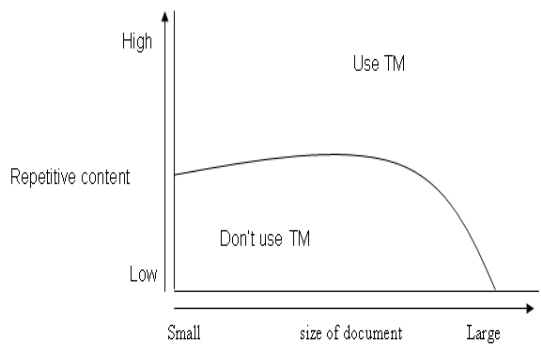

Figure1: Use of Translation Memory with respect to the size of document

## 2. TRANSLATION MEMORY STANDARDS

There are many standards available for creation, usage and data transfer of Translation Memories. Some of the prominent one are:

### 2.1 Translation Memory Exchange
Translation Memory Exchange (TMX) is an open XML standard which ensures the correct exchange of data between various translation memories. This is developed and maintained by Open Standards for Content Allowing Reuse (OSCAR) of Localization Industry Standards Association (LISA). It has been in existence since 1998. The exchange facilitates transfer of TM data among tools and translators with minimal or no loss of critical data [2]. The latest version of the standard was introduced in 2007. The major benefits of the standard are maximum flexibility, future proofing against technology change, access to broader pool of service providers and above all control over TM assets.

### 2.2 Term Base Exchange
Term Base Exchange (TBX) is a standard which was developed by LISA in 2002, but from 2008 onwards, it has been incorporated by ISO as ISO30042 standard. This standard was created to represent and exchange terminological data [3]. With the use of this standard a translator can maintain better quality and flexibility for terminological exchange of data from one tool to other.

### 2.3 Segmentation Rules Exchange
Segmentation Rules Exchange (SRX) is a standard which was again developed by LISA. This standard specifies the way in which translated text should be segmented[4]. As different languages and linguistic theories segment data differently, there was a constant need to develop a standard which could standardize the segmentation procedure for different families of languages. This standard actually works in conjunction with the TMX standard.

### 2.4 Universal Terminology Exchange
Universal Terminology Exchange (UTX) is a standard being developed by Asia Pacific Machine Translation Association (AAMT). The standard was required because, at times it was felt that the standard being formulated by LISA are meant only for commercial vendors of translation memories. This new standard will be applicable on all types of translations [5]. Moreover these translations can be further used in other NLP tasks like text to speech synthesis, automatic dictionary creation, thesaurus, input methods etc. This standard is also XML based standard, that is, this standard too advocates data to be available in XML format, so that better reusability can be provided, which eventually reduces cost and optimizes performance.

## 3. PROPOSED WORK

A lot of translation memories have been developed for European or East Asian Languages. But, there has been no effort on the Indian front. Then too mostly all the translation memories search text on complete sentences and cannot produce text in target language, if it is not available in translation database. Although, most TM tools claim to perform fuzzy matches, they too could not perform well as their match score is too low and other sentence are given preference over them. Moreover, they too fail to produce text for out of vocabulary sentences. As suggested by Macklovitch and Russell [6], the search in translation database should be treated as an Information Retrieval (IR) activity, as a TM tool only searches a large source of text documents for a desired match of a sentence in source language. We have tried to resolve this issue by using some statistical techniques. We have developed a translation memory which can help translators in providing better translations among Indian languages. At present we have incorporated English and Hindi as the language for our first stage of development and evaluation.

## 3.1 METHODOLOGY

To start with we required bilingual aligned corpus. We developed the corpus as described by Dash [7]. Moreover we employed the search procedure using n-gram approach as discussed by Jurafsky and Martin [8]. This approach is far more effective then direct or fuzzy searches employed by TM tools. In this approach, we orthographically related phrases, which had similar pairs.

| Sentences | Trigram Sequences |
|---|---|
| Café Coffee Day | Café Coffee Day |
| It has excellent menu and service | It has excellent, has excellent menu, excellent menu and, menu and service |
| Coffee at Café Coffee Day is good | Coffee at Café, at Café Coffee, Café Coffee Day, Coffee Day is, Day is good |
| Café Coffee Day has excellent menu and service | Café Coffee Day, Coffee Day has, Day has excellent, has excellent menu, excellent menu and, menu and service |

Table 1: Trigram Example

While determining whether two phrases were similar or not, we only needed to compare their words, as we used a three-gram or trigram approach. Here is an example to explain the same. Let us suppose, we have three sentences available in our translation database, which were selected for translation. These are: Sentence 1 as "Café Coffee Day", Sentence 2 as "It has excellent menu and service" and Sentence 3 as "Coffee at Café Coffee Day is good". We also have a Target Sentence as "Café Coffee Day has excellent menu and service". We break these entire sentences into trigrams as shown in Table 1.

We measure similarity score using Dice's Coefficient [9].

Sentence 1 – Target Sentence Pair = $3 \times \frac{1}{1+6}$ = 0.43

Sentence 2 – Target Sentence Pair = $3 \times \frac{3}{4+6}$ = 0.9

Sentence 3 – Target Sentence Pair = $3 \times \frac{1}{5+6}$ = 0.27

---

**Chunks:**

Will [VP they recommend] [NP our proposal] [VP made for sites]?

**Candidate Matches:**

<will they recommend>, <they recommend>, <they recommend our>, <recommend our>, <they recommend our proposal>, <our proposal>, <our proposal made >, <our proposal made for>, <our proposal made for sites>, <proposal made for>, <made for>, <made for sites>

**Actual Matches:**

<they recommend>, <they recommend our proposal>, <our proposal>

---

Figure2: Chunking sequences and matching phrases

So, as per this approach, we can decipher that Sentence 2 is the most appropriate sentence, among the list of candidate sentences and thus it should be translated. But, again this approach is not sufficient; this method too, cannot handle sentences which are not available in our database. So, we moved on to perform some linguistic analysis onto the sentences. With the help of a text chunker, we divided the entire sentence into chunks and got simple surface syntactic structures. Source sentences in the translation database were also divided into phrases and were matched with the chunks produced. Only the ones whose beginning or end was matched were retained for ranking and target translation. Figure 2 provides a brief example.

Here, we have found some matching phrases, we can again apply n-gram technique and can rank the chunks. We can provide the list of suggestive translations to the translator for the chunks which were matched. Then again this can create a problem, as if a chunk is too common, the translator will be flooded with translations, therefore we restricted our approach and selected the chunks which had maximum matching words. Adding this constraint reduces the burden on the part of the translator and is more intuitive as it follows the general perception that larger chunks or matched phrases are more helpful for the translator. Here, we not only provide the matched sentence, but also provide the list of top 5 suggestive translations, for the translator to judge, which is more correct.

## 4. EVALUATION

We developed a Bilingual text of 10,000 sentences for our translation database. We tested our system for three sets of 200 sentences each. The first set of 200 sentences were complete matching sentences available in the translation database. The second set had partial matching sentences and the third set had out of vocabulary sentences, the ones which were not available in the database. We compared the performance of our system with a human translator who manually translated these sentences. To provide a level playing field to the human translator, we provided the entire translation database to the human translator for reference.

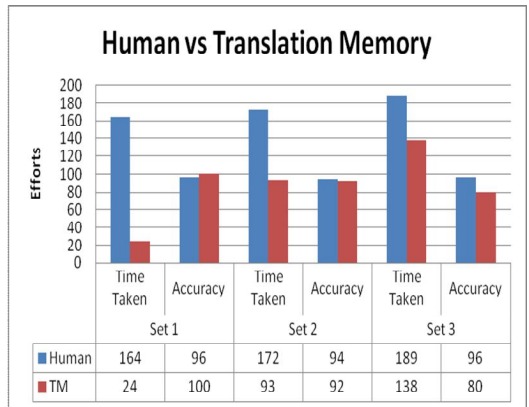

Figure2: Translation Memory vs Manual Translation

Evaluation was done on the basis of the correct translation done in the less amount of time and with less effort. In the first set, the TM tool provided 100% accurate result in 25 minutes approx. whereas the human translator took almost three hours to complete the process. In the second set, the TM tool provided partial translations for the sentences, which were then completed by a human translator. The total time taken to complete the process was one and a half hour approx. The manual translation again took the same amount of time i.e approximately three hours. In the third set of sentences, the TM tool could provide only partial translations for 68 sentences. The rest were left un-translated. Here the human effort to translate the remaining sentences was more. The entire translation process took two hours and fifteen minutes to complete. The manual translation again tool the same amount of time. In all the three cases the human accuracy was near to 100% whereas in first document the TMs accuracy was 100%, in second it was 92% and in the third it was reduced to 86%. Figure 3 summarizes the entire process.

## 5. CONCLUSION AND FUTURE WORK

We gave an outline for a Translation Memory Tool, which could facilitate the work of translators, who either have to post edit machine translations or have to provide the entire translations. Our objective was to provide a system where a human translator could provide better translations with minimum efforts. We have achieved this to some extent as the human translator can not only get translations for sentences or phrases, which are directly available in the translation database, but also can provide suggestive translations for sentences or phrases which are not readily available in the database. Since this system is still in its alpha phase, an immediate future study could be a more efficient and robust evaluation of its performance. A long term goal is to implement the system for all 22 constitutional languages of India. Then another enhancement to this system could be to implement a more optimized method like HMM or Maximum Entropy Model, for getting better target translations.